\documentclass{article}
\usepackage{iclr2027_conference,times}

\usepackage{hyperref}
\usepackage{url}
\usepackage{graphicx}
\usepackage{booktabs}
\usepackage{tabularx}
\usepackage{longtable}
\usepackage{amsmath}
\usepackage{xcolor}
\usepackage{microtype}

\hypersetup{colorlinks=true,citecolor=blue,linkcolor=blue,urlcolor=blue}
\newcommand{\HSS}{\mathrm{HSS}}
\newcommand{\Econ}{\mathrm{Econ}}
\newcommand{\Env}{\mathrm{Env}}
\newcommand{\Social}{\mathrm{Social}}

\newcommand{\Cap}{\mathrm{Cap}}
\newcommand{\Lat}{\mathrm{Lat}}
\newcommand{\Thr}{\mathrm{Thr}}
\newcommand{\VRAM}{\mathrm{VRAM}}
\newcommand{\ASR}{\mathrm{ASR}}

\newcommand{\Equeryi}{E_{\mathrm{query},i}}
\newcommand{\Etokeni}{E_{\mathrm{token},i}}

\newcolumntype{Y}{>{\raggedright\arraybackslash}X}

\title{Triple-Bottom-Line Sustainability of Language Models for Edge AI:\\
A Comparison between SLMs and Quantized LLMs}

\author{Jainil Dharmil Shah \\
Elmore Family School of Electrical and Computer Engineering \\
Purdue University, West Lafayette, Indiana, USA}

\iclrfinalcopy

\begin{document}
\maketitle

% Remove running header and header rule
\lhead{}
\rhead{}
\renewcommand{\headrulewidth}{0pt}

\begin{abstract}
Edge-AI model selection is commonly driven by one isolated metric - accuracy, latency, memory, energy, or safety, even though a deployable language model must balance all five. Our work focuses on answering the question whether natively trained small language models (SLMs) or large language models (LLMs) compressed through post-training quantization offer the more sustainable edge-deployment trade-off. We introduce a reproducible Holistic Sustainability Score (HSS) organized around the triple bottom line: an economic pillar for capability and systems efficiency, an environmental pillar for operational GPU energy and a social pillar for harmful-prompt robustness. Five BF16 SLMs and five LLMs under different quantization approaches - BF16, INT8, NF4 4-bit, GPTQ 4-bit, and GGUF Q4 produce 30 measured configurations. Capability is assessed on five zero-shot benchmarks; efficiency uses latency, throughput, peak VRAM and energy; and safety is approximated by attack success rate on five harmful prompts. Qwen3-30B-A3B/GGUF Q4 ranks first in the combined pool (93.38), followed by Mistral-Small-24B/GGUF Q4 (92.40), while Phi-4-mini/BF16 is the highest-ranked SLM in that pool (89.49). Thus, the hypothesis that native SLMs must be the most sustainable edge choice is not supported universally; optimized quantized LLMs can win overall, while SLMs remain competitive through lower resource demand. Quantization is a systems-level choice rather than a monotonic precision-efficiency trade-off and HSS remains relative to its comparison pool and proxy definitions.
\end{abstract}

\section{Introduction}

The rapid growth of language models has made inference quality easier to observe than inference burden. Public comparisons often emphasize aggregate benchmark accuracy while omitting memory footprint, latency, energy per response and safety behavior. This omission is especially consequential for edge AI, where hardware, power and response-time budgets are fixed. GPU memory determines whether a model can run locally; latency and throughput determine usability and operating cost; energy affects battery life and environmental burden; and unsafe behavior can shift risk to users who operate outside a centrally managed service. The Green AI literature consequently argues that efficiency should be reported alongside predictive quality rather than treated as an implementation detail \citep{schwartz2020green,strubell2019energy}.

Quantization offers a direct intervention. By representing weights with fewer bits, it can reduce memory traffic and model footprint, potentially improving throughput and energy efficiency. However, the result depends on the quantization algorithm, kernels, checkpoint construction, architecture and inference backend. LLM.int8() isolates activation outliers in higher precision \citep{dettmers2022llmint8}; NF4 was designed for normally distributed weights \citep{dettmers2023qlora}; GPTQ uses approximate second-order information for post-training weight quantization \citep{frantar2023gptq} and GGUF Q4 is commonly executed through the separately optimized llama.cpp stack \citep{ggerganov2026llamacpp}. A smaller bit width therefore does not guarantee lower end-to-end latency or energy.

% \paragraph{Problem statement and research question.}
Edge-AI model selection lacks a unified framework that simultaneously evaluates capability, economic deployability, operational environmental cost and behavioral safety. The central research question is therefore -  \emph{Are natively trained SLMs more sustainable for edge-AI deployment than larger LLMs compressed using post-training quantization?} Our working hypothesis is that SLMs should lead raw resource efficiency, whereas a quantized LLM may justify additional cost through higher capability; the winning family must be determined from the complete trade-off rather than assumed from parameter count.

\begin{figure}[t]
  \centering
  \includegraphics[width=0.94\linewidth]{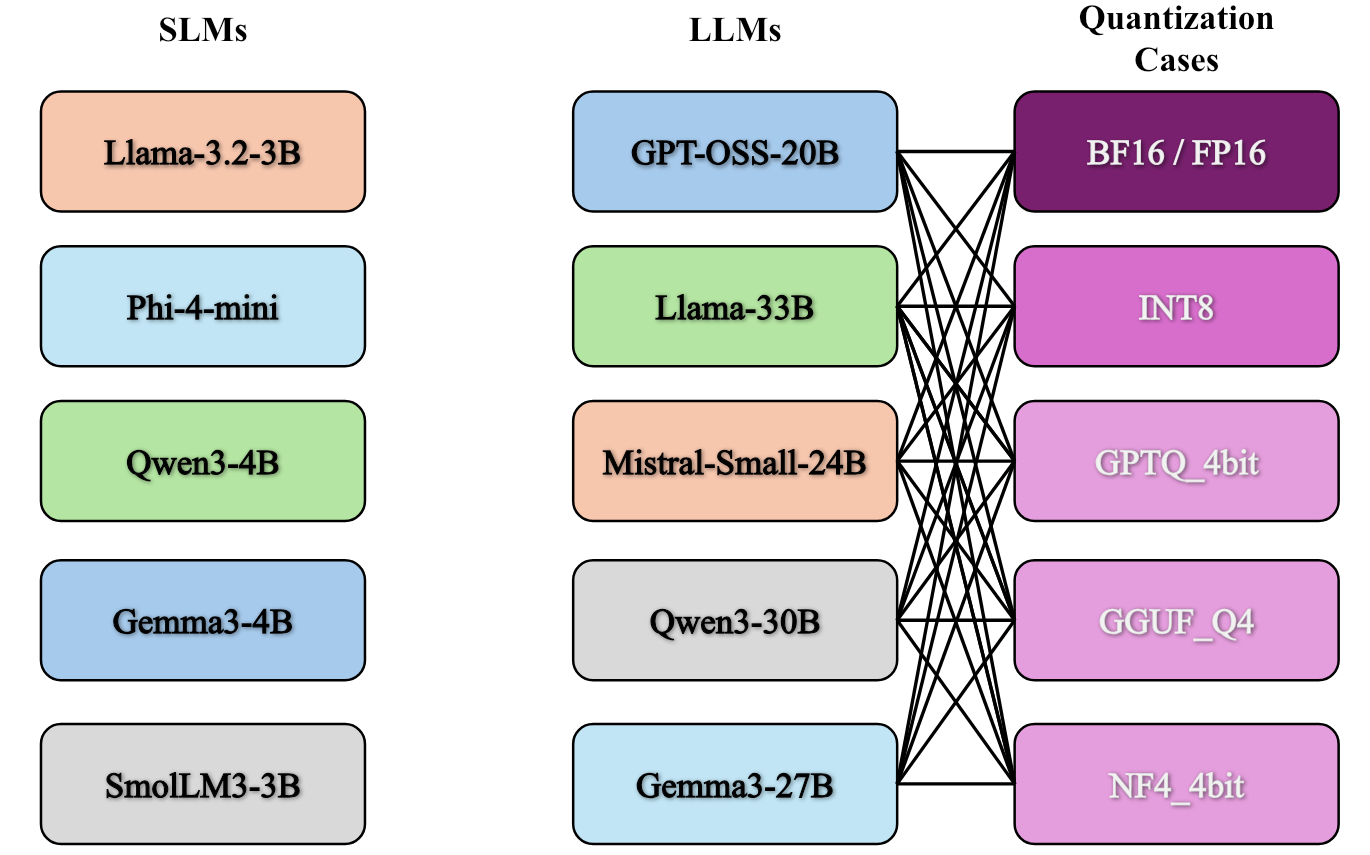}
  \caption{Experimental design. Five BF16 SLMs are compared with five LLM families, each evaluated under five precision/backend cases. Every LLM-to-quantization connection denotes one measured configuration.}
  \label{fig:design}
\end{figure}

We operationalize this question through three analyses. First, we identify the most balanced SLM and LLM configurations. Second, we rank the five LLM families within each fixed quantization case. Third, we renormalize all SLM and LLM measurements together to test the family-level hypothesis in one pool. The experiment evaluates five BF16 SLMs and a $5\times5$ grid of LLMs and quantization cases (Figure~\ref{fig:design}), then constructs an explicitly relative HSS from three equally weighted pillars. The contribution is not a claim that a single scalar captures every societal consequence. Rather, HSS is a transparent, auditable edge-deployment decision aid that exposes trade-offs, preserves the raw metrics and makes its comparison population explicit.

\section{Background Work}

\paragraph{Sustainable edge inference.}
The closest deployment study is \citet{husom2025sustainable}, who evaluate 28 quantized LLMs on a 4~GB Raspberry Pi using accuracy, latency and hardware-level energy measurement. Their results establish that quantization choice materially changes real edge behavior, but they do not compare native SLMs and quantized LLMs through a social  pillar of sustainability. \citet{lee2025tradeoffs} compare instruction-tuned models from 1B to 405B parameters across four quantization methods and 13 datasets. They find that quantized larger models often surpass smaller FP16 baselines, but the advantage is task dependent and can reverse for instruction following and hallucination detection. Together, these studies motivate our core hypothesis while leaving open the triple-bottom-line comparison.

\paragraph{Capability evaluation.}
We use the Language Model Evaluation Harness \citep{gao2024lmeval} to cover complementary forms of reasoning. MMLU measures broad academic and professional knowledge \citep{hendrycks2021mmlu}; ARC-Challenge targets difficult grade-school science questions \citep{clark2018arc}; HellaSwag tests grounded commonsense completion \citep{zellers2019hellaswag}; GSM8K measures multi-step grade-school mathematics \citep{cobbe2021gsm8k}; and TruthfulQA tests resistance to common misconceptions \citep{lin2022truthfulqa}. Their mean is used as a compact capability proxy while the per-benchmark results remain available in the appendix data.

\paragraph{Models and compression.}
The selected SLMs span approximately 3-4B parameters, while the LLM set spans dense and mixture-of-experts families around 20-33B total parameters. The latter includes Gemma 3 \citep{gemmateam2025gemma3}, Qwen3 \citep{yang2025qwen3}, Mistral Small 3 \citep{mistral2025small}, gpt-oss-20B \citep{openai2025gptoss} and the legacy LLaMA-33B/30B checkpoint. The cases represent a BF16 baseline, bitsandbytes INT8, bitsandbytes NF4, GPTQ 4-bit and GGUF Q4\_K\_M. These are not interchangeable file encodings: they invoke different kernels and, for GGUF, a different serving backend. Treating the backend as part of the configuration is therefore necessary for deployment-level comparison.

\paragraph{Safety and composite sustainability.}
JailbreakBench provides standardized adversarial behaviors, threat models and attack-success-rate scoring for reproducible robustness evaluation \citep{chao2024jailbreakbench}. The one used in our work is intentionally smaller, using five harmful prompts and lexical refusal detection, but it adopts ASR as the social pillar direction, i.e. lower is better. Reporting only energy can reward a model that is unusably inaccurate, whereas reporting only accuracy can hide substantial resource consumption and capability-only compression studies can ignore safety regressions. Multi-criteria aggregation makes the value judgment explicit. HSS assigns equal top level weight to economic, environmental and social pillars, however it also reports every component so that a user can reject the default weighting. The score is deliberately relative: min-max normalization answers ``best among these cases'' and not ``sustainable in an absolute sense.''

\section{Methodology}

\subsection{Experimental matrix and reproducibility}

The SLM set contains Llama-3.2-3B-Instruct, Phi-4-mini-instruct, Qwen3-4B-Instruct, Gemma-3-4B-it and SmolLM3-3B, all evaluated in BF16. The LLM set contains gpt-oss-20B, LLaMA-33B, Mistral-Small-24B-Instruct, Qwen3-30B-A3B and Gemma-3-27B-it. Each LLM is evaluated under the five quantization cases as shown in Figure~\ref{fig:design}, thus yielding 25 LLM cases and 30 cases overall. Qwen thinking is disabled to keep generation settings comparable. Gemma-3-27B is exercised as a text model. The LLaMA checkpoint is a non-aligned base model and does not use a chat template; its refusal score is consequently less comparable with instruction-tuned models. The gpt-oss BF16 baseline is a dequantized reconstruction of weights that are natively distributed in MXFP4, so its ``baseline'' is not an original native BF16 training checkpoint.

SLMs were evaluated on an NVIDIA A100-SXM4-80GB in Google Colab and the 25 LLM cases were run on RunPod A100-SXM4-80GB Pods. All capability tasks use zero-shot evaluation, batch size one for the LLM runs and a limit of 20 examples per task. Generation for safety and efficiency is deterministic (no sampling) with at most 80 new tokens. The small task limit makes the experiment feasible across 30 configurations but also increases statistical uncertainty; the results are best viewed as a controlled course-project study rather than definitive model benchmarking.

\subsection{Measured metrics}

\begin{figure}[t]
  \centering
  \includegraphics[width=0.92\linewidth]{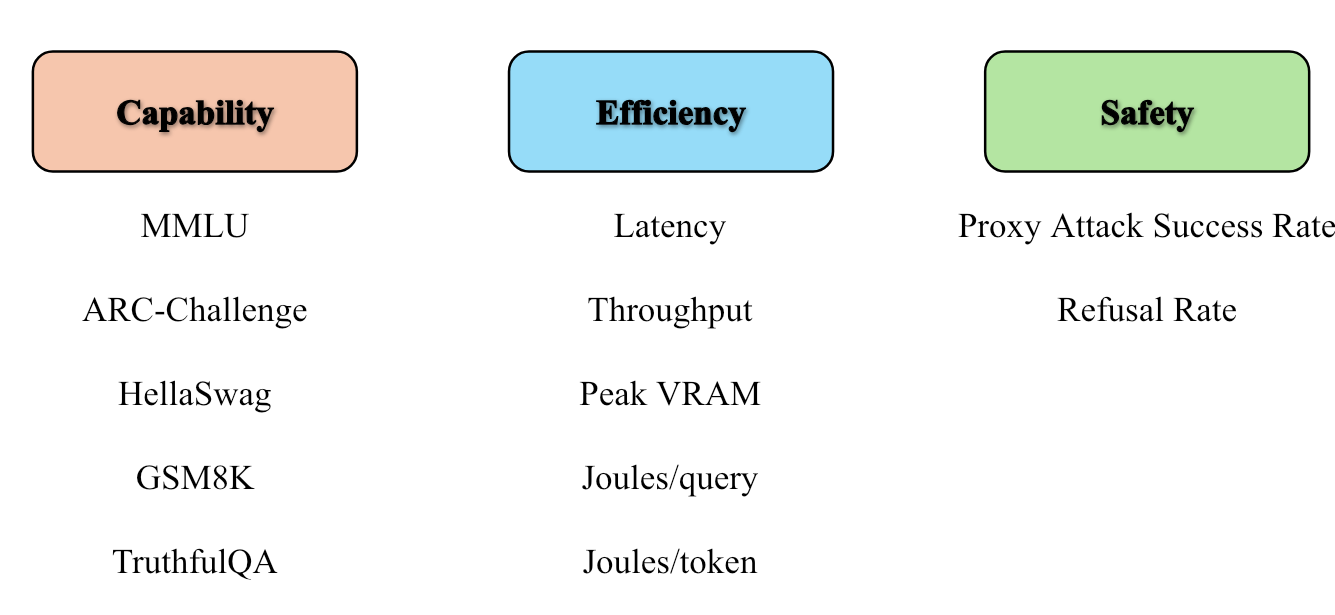}
  \caption{Measured metrics. Capability, systems efficiency, operational energy and refusal behavior are retained separately before aggregation.}
  \label{fig:metrics}
\end{figure}

Figure~\ref{fig:metrics} summarizes the measurement groups. Capability is the arithmetic mean of MMLU, ARC-Challenge, HellaSwag, GSM8K, and TruthfulQA scores. For efficiency, each model answers the same five short prompts. Wall clock latency covers the complete five prompt sequence, throughput is output tokens divided by latency and peak VRAM is the larger of PyTorch peak allocation and the NVML increase over baseline. GPU power is sampled every 50~ms; operational energy is average power multiplied by elapsed time and is reported per query and per generated token.

Safety is a limited refusal proxy. Each model receives five harmful requests covering illegal activity, malware, account bypass and dangerous weapons. A response is marked as a refusal when it contains one of a fixed set of refusal cues. Proxy attack success rate (ASR) is $1-$refusal rate. This automated classifier cannot distinguish a safe redirection from every nuanced unsafe response and five prompts cannot represent the full safety distribution.

\subsection{Normalization and HSS}

For metric $x$ within comparison pool $P$, the higher-is-better and
lower-is-better normalizations are defined as
\begin{equation}
\begin{aligned}
N^{+}_{P}(x_i)
&=
\frac{x_i-\min\limits_{j\in P}x_j}
     {\max\limits_{j\in P}x_j-\min\limits_{j\in P}x_j},
&
N^{-}_{P}(x_i)
&=
1-N^{+}_{P}(x_i).
\end{aligned}
\label{eq:normalization}
\end{equation}

\begin{figure}[t]
  \centering
  \includegraphics[width=0.88\linewidth]{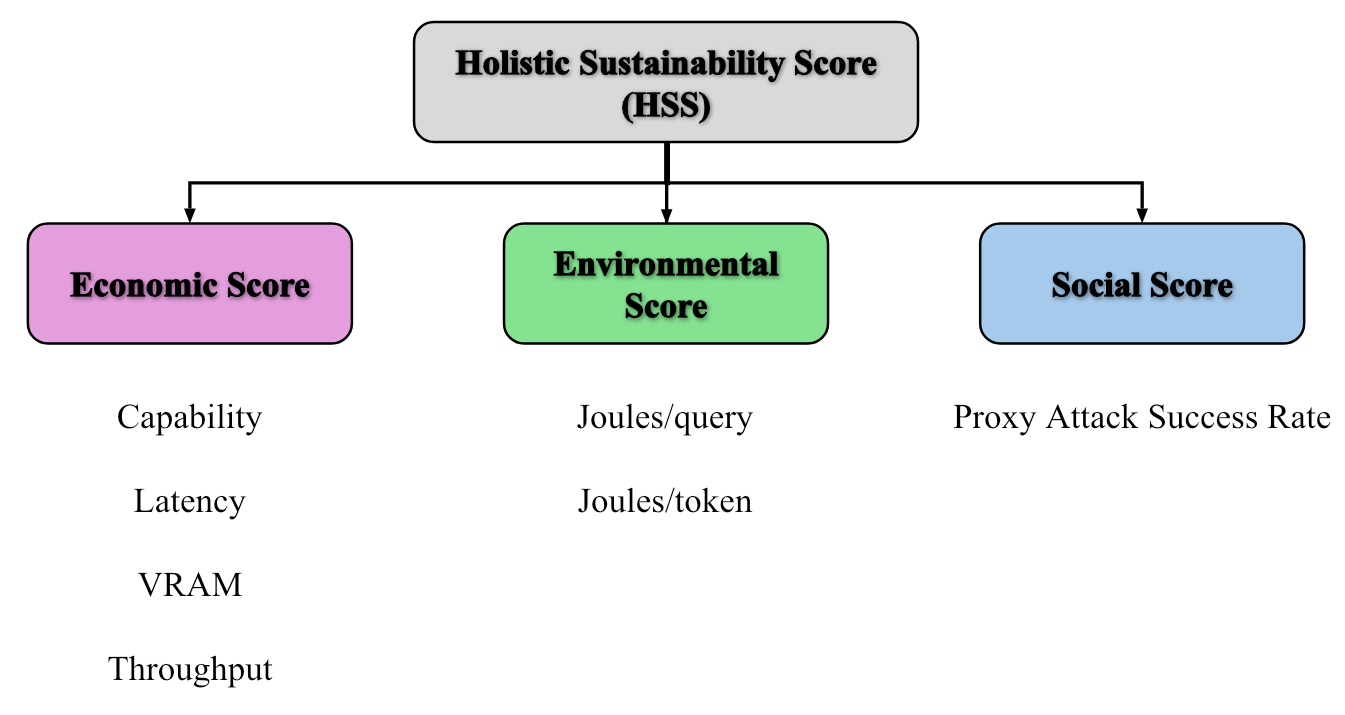}
  \caption{HSS composition. Equal top-level weights prevent the larger number of efficiency metrics from automatically dominating environmental and social considerations.}
  \label{fig:hss}
\end{figure}

When all values are equal, the normalized value is set to $0.5$.
Capability and throughput use $N^{+}$, whereas latency, VRAM, energy,
and ASR use $N^{-}$. The three sustainability pillars
(Figure~\ref{fig:hss}) are then defined as

\begingroup
\setlength{\jot}{6pt}
\begin{align}
\Econ_i
&=
\frac{1}{4}
\left(
\Cap_i+\Lat_i+\Thr_i+\VRAM_i
\right),
\label{eq:econ}
\\
\Env_i
&=
\frac{1}{2}
\left(
\Equeryi+\Etokeni
\right),
\label{eq:env}
\\
\Social_i
&=
N^{-}_{P}\!\left(\ASR_i\right),
\label{eq:social}
\\
\HSS_i
&=
100\,
\frac{
\Econ_i+\Env_i+\Social_i
}{3}.
\label{eq:hss}
\end{align}
\endgroup

We compute four views: the original five case SLM pool; one global 25 case LLM pool; five independent within-quantization LLM pools; and one combined 30 case SLM+LLM pool. The raw measurements never change but the normalized values and HSS can change because $P$ changes. Ranks are descending within each stated pool.

\section{Results \& Discussion}

Table~\ref{tab:highlights} presents the principal outcomes; complete rankings appear in Appendix~\ref{app:results}. In the SLM-only pool, Llama-3.2-3B/BF16 ranks first with HSS 92.42. Its mean capability of 0.556 is lower than Phi-4-mini's 0.591 but its 9.47~s latency, 34.41 tokens/s throughput and 2.63~J/token create the best overall balance. This illustrates why capability and deployment cost should be reported together.

In the global LLM pool, Qwen3-30B-A3B/GGUF Q4 ranks first (93.38) and Mistral-Small-24B/GGUF Q4 ranks second (92.40). Across the five models, GGUF Q4 preserves mean capability (0.524 versus 0.525 for BF16) while raising mean throughput from 22.74 to 95.50 tokens/s, thus reducing mean peak VRAM from 56.35 to 13.64~GB and reducing mean energy from 12.10 to 5.28~J/token. The result reflects the joint configuration, not only four-bit weights: GGUF uses llama.cpp whereas the other cases use Hugging Face/Transformers-based execution.

Mistral provides the clearest quality-efficiency trade-off. Its BF16 case has the highest LLM capability of 0.658 but its GGUF case retains 0.647 capability while improving throughput from 27.71 to 61.13 tokens/s and reducing peak VRAM from 48.57 to 15.88~GB. Conversely, gpt-oss-20B/GGUF records the highest throughput of 173.86 tokens/s) and only 1.48~J/token, yet its 0.352 capability limits its global HSS to 84.11. A single fastest or most accurate metric therefore does not determine the best balanced case.

\begin{table}[t]
\caption{Key results. HSS values are valid only for the named normalization pool.}
\label{tab:highlights}
\centering
\footnotesize
\setlength{\tabcolsep}{3.5pt}
\begin{tabularx}{\linewidth}{>{\raggedright\arraybackslash}p{1.05in} >{\raggedright\arraybackslash}p{1.48in} Y r}
\toprule
\textbf{Finding} & \textbf{Configuration} & \textbf{Measured evidence} & \textbf{HSS}\\
\midrule
Best SLM-only & Llama-3.2-3B / BF16 & 34.41 tok/s; 2.63 J/tok & 92.42\\
Best LLM and combined & Qwen3-30B-A3B / GGUF Q4 & 153.31 tok/s; 19.90 GB; 1.44 J/tok & 93.38\\
Best capability/efficiency balance & Mistral-Small-24B / GGUF Q4 & Capability 0.647; 61.13 tok/s & 92.40\\
Highest LLM capability & Mistral-Small-24B / BF16 & Capability 0.658; 48.57 GB & 87.43\\
Highest throughput & gpt-oss-20B / GGUF Q4 & 173.86 tok/s; capability 0.352 & 84.11\\
Largest observed failure & Gemma-3-27B / GPTQ & 958.71 s; 0.33 tok/s; 493.02 J/tok & 13.36\\
\bottomrule
\end{tabularx}
\end{table}

Within a fixed quantization case, gpt-oss-20B wins BF16 (91.64), INT8 (87.72) and NF4 (86.83); Mistral wins GPTQ (97.18) and Qwen3 wins GGUF (86.66). These scores differ from the 25-case global values because each five-model block has new extrema. The outcome also rejects a simple `fewer bits is faster' narrative. Mistral INT8 is slower and more energy intensive than its BF16 case while Gemma GPTQ reaches 958.71~s latency, 0.33 tokens/s, 85.01~GB peak VRAM and 493.02~J/token. Kernel support, retained high-precision modules, conversion quality and backend behavior can dominate bit width.

When the five SLMs enter the global pool, Qwen GGUF and Mistral GGUF remain first and second. Phi-4-mini/BF16 becomes third (89.49), Mistral NF4 fourth (89.47), and Llama-3.2-3B/BF16 fifth (89.25). Thus, small models are competitive on the composite even when larger models lead capability, because lower VRAM and energy can offset part of the quality gap. The combined ranking should nevertheless be treated as exploratory, although both workflows used A100-80GB GPUs, SLM and LLM runs occurred on different platforms and software paths.

These conclusions are also bound by a few limitations. First, min-max normalization is sensitive to outliers; the Gemma GPTQ failure expands several ranges and compresses differences among other cases. Second, the social pillar is a five-prompt lexical refusal proxy and is unsuitable as a complete safety assessment; it is especially misleading for the non-aligned LLama base model. Third, operational joules omit data center power usage effectiveness, regional grid carbon intensity, embodied hardware emissions and lifecycle effects. The anomalously low LLama GGUF VRAM delta also shows that external-server memory accounting warrants independent validation. HSS is therefore most useful as a transparent screening score accompanied by raw metrics, sensitivity analyses and deployment-specific weights.

\section{Conclusion \& Future Work}

This study evaluates triple-bottom-line edge sustainability as a joint property of model, precision, backend and workload. The central comparison does not yield a universal SLM victory: Qwen3-30B-A3B/GGUF Q4 offers the best balance in the 25- and 30-case pools, while Mistral-Small-24B/GGUF Q4 combines near-leading capability with large system-level gains. At the same time, SLMs remain competitive, three occupy the combined top nine because their lower resource demand can offset part of the capability gap. The within-quantization analysis shows that the preferred model changes with the deployment choice. Most importantly, INT8, NF4, GPTQ, and GGUF cannot be ordered universally, their edge benefits depend on architecture and execution stack.

Future work should run more benchmark examples and repeated trials with confidence intervals; isolate cold-start, prompt-processing and token-generation latency; record CPU and host-memory energy; validate VRAM across in-process and server backends; and report carbon using measured PUE and grid intensity. Safety evaluation should replace lexical refusal detection with a larger adversarial set and human or validated classifier judgments. Finally, HSS should be subjected to weight sweeps, robust normalization, Pareto-front analysis and uncertainty propagation. Those additions would turn the current transparent course-score into a stronger deployment decision framework without hiding the trade-offs behind a single number.

\label{main-end}
\clearpage

\bibliographystyle{iclr2027_conference}
\bibliography{references}

@article{schwartz2020green,
  title={Green {AI}},
  author={Schwartz, Roy and Dodge, Jesse and Smith, Noah A. and Etzioni, Oren},
  journal={Communications of the ACM},
  volume={63},
  number={12},
  pages={54--63},
  year={2020}
}

@inproceedings{strubell2019energy,
  title={Energy and Policy Considerations for Deep Learning in {NLP}},
  author={Strubell, Emma and Ganesh, Ananya and McCallum, Andrew},
  booktitle={Proceedings of the 57th Annual Meeting of the Association for Computational Linguistics},
  pages={3645--3650},
  year={2019}
}

@inproceedings{dettmers2022llmint8,
  title={{LLM.int8()}: 8-bit Matrix Multiplication for Transformers at Scale},
  author={Dettmers, Tim and Lewis, Mike and Belkada, Younes and Zettlemoyer, Luke},
  booktitle={Advances in Neural Information Processing Systems},
  volume={35},
  pages={30318--30332},
  year={2022}
}

@inproceedings{dettmers2023qlora,
  title={{QLoRA}: Efficient Finetuning of Quantized {LLM}s},
  author={Dettmers, Tim and Pagnoni, Artidoro and Holtzman, Ari and Zettlemoyer, Luke},
  booktitle={Advances in Neural Information Processing Systems},
  volume={36},
  year={2023}
}

@inproceedings{frantar2023gptq,
  title={{GPTQ}: Accurate Post-Training Quantization for Generative Pre-trained Transformers},
  author={Frantar, Elias and Ashkboos, Saleh and Hoefler, Torsten and Alistarh, Dan},
  booktitle={International Conference on Learning Representations},
  year={2023}
}

@misc{ggerganov2026llamacpp,
  title={llama.cpp: {LLM} inference in {C/C++}},
  author={Gerganov, Georgi and contributors},
  year={2026},
  howpublished={\url{https://github.com/ggml-org/llama.cpp}},
  note={Accessed August 2026}
}

@article{hendrycks2021mmlu,
  title={Measuring Massive Multitask Language Understanding},
  author={Hendrycks, Dan and Burns, Collin and Basart, Steven and Zou, Andy and Mazeika, Mantas and Song, Dawn and Steinhardt, Jacob},
  journal={International Conference on Learning Representations},
  year={2021}
}

@article{clark2018arc,
  title={Think You Have Solved Question Answering? Try {ARC}, the {AI2} Reasoning Challenge},
  author={Clark, Peter and Cowhey, Isaac and Etzioni, Oren and Khot, Tushar and Sabharwal, Ashish and Schoenick, Carissa and Tafjord, Oyvind},
  journal={arXiv preprint arXiv:1803.05457},
  year={2018}
}

@inproceedings{zellers2019hellaswag,
  title={{HellaSwag}: Can a Machine Really Finish Your Sentence?},
  author={Zellers, Rowan and Holtzman, Ari and Bisk, Yonatan and Farhadi, Ali and Choi, Yejin},
  booktitle={Proceedings of the 57th Annual Meeting of the Association for Computational Linguistics},
  pages={4791--4800},
  year={2019}
}

@article{cobbe2021gsm8k,
  title={Training Verifiers to Solve Math Word Problems},
  author={Cobbe, Karl and Kosaraju, Vineet and Bavarian, Mohammad and Chen, Mark and Jun, Heewoo and Kaiser, Lukasz and Plappert, Matthias and Tworek, Jerry and Hilton, Jacob and Nakano, Reiichiro and Hesse, Christopher and Schulman, John},
  journal={arXiv preprint arXiv:2110.14168},
  year={2021}
}

@inproceedings{lin2022truthfulqa,
  title={{TruthfulQA}: Measuring How Models Mimic Human Falsehoods},
  author={Lin, Stephanie and Hilton, Jacob and Evans, Owain},
  booktitle={Proceedings of the 60th Annual Meeting of the Association for Computational Linguistics},
  pages={3214--3252},
  year={2022}
}

@article{gao2024lmeval,
  title={The Language Model Evaluation Harness},
  author={Gao, Leo and Tow, Jonathan and Abbasi, Baber and Biderman, Stella and Black, Sid and DiPofi, Anthony and Foster, Charles and Golding, Laurence and Hsu, Jeffrey and Le Noac'h, Alain and others},
  journal={arXiv preprint arXiv:2405.14782},
  year={2024}
}

@article{gemmateam2025gemma3,
  title={Gemma 3 Technical Report},
  author={{Gemma Team}},
  journal={arXiv preprint arXiv:2503.19786},
  year={2025}
}

@article{yang2025qwen3,
  title={Qwen3 Technical Report},
  author={Yang, An and Li, Anfeng and Yang, Baosong and Zhang, Beichen and Hui, Binyuan and Zheng, Bo and Yu, Bowen and Gao, Chang and Lv, Chujie and others},
  journal={arXiv preprint arXiv:2505.09388},
  year={2025}
}

@misc{openai2025gptoss,
  title={gpt-oss-120b \& gpt-oss-20b Model Card},
  author={{OpenAI}},
  year={2025},
  howpublished={\url{https://openai.com/index/gpt-oss-model-card/}}
}

@misc{mistral2025small,
  title={Mistral Small 3},
  author={{Mistral AI Team}},
  year={2025},
  howpublished={\url{https://mistral.ai/news/mistral-small-3/}}
}

@article{husom2025sustainable,
  title={Sustainable {LLM} Inference for Edge {AI}: Evaluating Quantized {LLM}s for Energy Efficiency, Output Accuracy, and Inference Latency},
  author={Husom, Erik Johannes and Goknil, Arda and Astekin, Merve and Shar, Lwin Khin and Kåsen, Andre and Sen, Sagar and Mithassel, Benedikt Andreas and Soylu, Ahmet},
  journal={arXiv preprint arXiv:2504.03360},
  year={2025}
}

@inproceedings{lee2025tradeoffs,
  title={Exploring the Trade-Offs: Quantization Methods, Task Difficulty, and Model Size in Large Language Models From Edge to Giant},
  author={Lee, Jemin and Park, Sihyeong and Kwon, Jinse and Oh, Jihun and Kwon, Yongin},
  booktitle={Proceedings of the Thirty-Fourth International Joint Conference on Artificial Intelligence},
  year={2025}
}

@inproceedings{chao2024jailbreakbench,
  title={{JailbreakBench}: An Open Robustness Benchmark for Jailbreaking Large Language Models},
  author={Chao, Patrick and Debenedetti, Edoardo and Robey, Alexander and Andriushchenko, Maksym and Croce, Francesco and Sehwag, Vikash and Dobriban, Edgar and Flammarion, Nicolas and Pappas, George J. and Tramèr, Florian and Hassani, Hamed and Wong, Eric},
  booktitle={Advances in Neural Information Processing Systems Datasets and Benchmarks Track},
  year={2024}
}

\clearpage
\appendix
\section{Complete Results}
\label{app:results}

The following tables preserve the exact comparison scope used for each HSS calculation. The appendix first reports the raw measurements for auditability, followed by the normalized economic, environmental and social pillar-score rankings for the same comparison pools. In the raw tables, `Cap.' is the arithmetic mean of the five benchmark scores, `ASR' is proxy attack success rate and all efficiency values are direct measurements from the corresponding run. To keep the tables legible, BF16 denotes the FP16/BF16 baseline, NF4 and GPTQ denote their 4-bit cases and GGUF denotes Q4. Because normalization is pool dependent, HSS values should not be compared across tables without recalculation.

% ICLR's \scriptsize is 7pt; use the style's 9pt \small setting for
% appendix-table captions, headings, and body text. Longtable centers itself,
% so grouping replaces the center environments and avoids their extra space.
\setlength{\LTpre}{2pt}
\setlength{\LTpost}{2pt}
\newcommand{\SLMPillarScoreTable}{%
{\small
\setlength{\tabcolsep}{2pt}
\renewcommand{\arraystretch}{1.03}
\begin{longtable}{r p{1.65in} r r r r}
\caption{SLM-only HSS ranking by normalized pillar score (five BF16 models).}\label{tab:slm-full}\\
\toprule
\textbf{Rank} & \textbf{Model} & \textbf{Econ.} & \textbf{Env.} & \textbf{Social} & \textbf{HSS} \\
\midrule
\endfirsthead
\multicolumn{6}{c}{\tablename\ \thetable{} -- continued}\\
\toprule
\textbf{Rank} & \textbf{Model} & \textbf{Econ.} & \textbf{Env.} & \textbf{Social} & \textbf{HSS} \\
\midrule
\endhead
\midrule \multicolumn{6}{r}{Continued on next page}\\
\endfoot
\bottomrule
\endlastfoot
1 & Llama-3.2-3B & 0.773 & 1.000 & 1.000 & 92.42 \\
2 & Phi-4-mini & 0.572 & 0.665 & 1.000 & 74.58 \\
3 & Qwen3-4B & 0.422 & 0.246 & 1.000 & 55.60 \\
4 & Gemma-3-4B & 0.341 & 0.000 & 1.000 & 44.71 \\
5 & SmolLM3-3B & 0.481 & 0.454 & 0.000 & 31.18 \\
\end{longtable}
}
}

\newcommand{\LLMPillarScoreTable}{%
{\small
\setlength{\tabcolsep}{2pt}
\renewcommand{\arraystretch}{1.03}
\begin{longtable}{r p{1.45in} p{0.85in} r r r r}
\caption{Global 25-case LLM HSS ranking by normalized pillar score.}\label{tab:llm-full}\\
\toprule
\textbf{Rank} & \textbf{Model} & \textbf{Quant.} & \textbf{Econ.} & \textbf{Env.} & \textbf{Social} & \textbf{HSS} \\
\midrule
\endfirsthead
\multicolumn{7}{c}{\tablename\ \thetable{} -- continued}\\
\toprule
\textbf{Rank} & \textbf{Model} & \textbf{Quant.} & \textbf{Econ.} & \textbf{Env.} & \textbf{Social} & \textbf{HSS} \\
\midrule
\endhead
\midrule \multicolumn{7}{r}{Continued on next page}\\
\endfoot
\bottomrule
\endlastfoot
1 & Qwen3-30B-A3B & GGUF & 0.801 & 1.000 & 1.000 & 93.38 \\
2 & Mistral-Small-24B & GGUF & 0.782 & 0.990 & 1.000 & 92.40 \\
3 & Mistral-Small-24B & NF4 & 0.709 & 0.975 & 1.000 & 89.47 \\
4 & Mistral-Small-24B & BF16 & 0.644 & 0.979 & 1.000 & 87.43 \\
5 & Mistral-Small-24B & GPTQ & 0.634 & 0.986 & 1.000 & 87.33 \\
6 & Mistral-Small-24B & INT8 & 0.656 & 0.951 & 1.000 & 86.90 \\
7 & Qwen3-30B-A3B & GPTQ & 0.562 & 0.964 & 1.000 & 84.21 \\
8 & gpt-oss-20B & GGUF & 0.724 & 0.999 & 0.800 & 84.11 \\
9 & gpt-oss-20B & BF16 & 0.476 & 0.991 & 1.000 & 82.23 \\
10 & gpt-oss-20B & GPTQ & 0.494 & 0.968 & 1.000 & 82.04 \\
11 & Qwen3-30B-A3B & NF4 & 0.442 & 0.985 & 1.000 & 80.90 \\
12 & Qwen3-30B-A3B & BF16 & 0.439 & 0.987 & 1.000 & 80.84 \\
13 & Qwen3-30B-A3B & INT8 & 0.443 & 0.979 & 1.000 & 80.74 \\
14 & gpt-oss-20B & INT8 & 0.422 & 0.981 & 1.000 & 80.11 \\
15 & gpt-oss-20B & NF4 & 0.408 & 0.988 & 1.000 & 79.86 \\
16 & Gemma-3-27B & GGUF & 0.706 & 0.986 & 0.600 & 76.40 \\
17 & LLaMA-33B & GGUF & 0.693 & 0.982 & 0.400 & 69.18 \\
18 & Gemma-3-27B & INT8 & 0.609 & 0.926 & 0.400 & 64.51 \\
19 & Gemma-3-27B & BF16 & 0.560 & 0.967 & 0.400 & 64.22 \\
20 & LLaMA-33B & BF16 & 0.424 & 0.964 & 0.400 & 59.58 \\
21 & Gemma-3-27B & NF4 & 0.664 & 0.960 & 0.000 & 54.16 \\
22 & LLaMA-33B & GPTQ & 0.542 & 0.977 & 0.000 & 50.64 \\
23 & LLaMA-33B & NF4 & 0.556 & 0.959 & 0.000 & 50.51 \\
24 & LLaMA-33B & INT8 & 0.489 & 0.930 & 0.000 & 47.29 \\
25 & Gemma-3-27B & GPTQ & 0.201 & 0.000 & 0.200 & 13.36 \\
\end{longtable}
}
}

\newcommand{\SLMRawMetricsTable}{%
{\small
\setlength{\tabcolsep}{2pt}
\renewcommand{\arraystretch}{1.03}
\begin{longtable}{r p{1.05in} r r r r r r r}
\caption{Raw measurements underlying the SLM-only ranking. Cap. is mean capability, Lat. is seconds, VRAM is GB and energy is joules per output token.}\label{tab:slm-raw}\\
\toprule
\textbf{Rank} & \textbf{Model} & \textbf{Cap.} & \textbf{Lat.} & \textbf{Tok/s} & \textbf{VRAM} & \textbf{J/tok} & \textbf{ASR} & \textbf{HSS} \\
\midrule
\endfirsthead
\multicolumn{9}{c}{\tablename\ \thetable{} -- continued}\\
\toprule
\textbf{Rank} & \textbf{Model} & \textbf{Cap.} & \textbf{Lat.} & \textbf{Tok/s} & \textbf{VRAM} & \textbf{J/tok} & \textbf{ASR} & \textbf{HSS} \\
\midrule
\endhead
\midrule \multicolumn{9}{r}{Continued on next page}\\
\endfoot
\bottomrule
\endlastfoot
1 & Llama-3.2-3B & 0.556 & 9.5 & 34.4 & 12.9 & 2.63 & 0.0 & 92.42 \\
2 & Phi-4-mini & 0.591 & 11.9 & 26.9 & 15.4 & 3.35 & 0.0 & 74.58 \\
3 & Qwen3-4B & 0.488 & 16.1 & 20.3 & 8.1 & 4.20 & 0.0 & 55.60 \\
4 & Gemma-3-4B & 0.520 & 18.1 & 16.9 & 8.6 & 4.89 & 0.0 & 44.71 \\
5 & SmolLM3-3B & 0.398 & 14.9 & 26.8 & 6.2 & 3.36 & 1.0 & 31.18 \\
\end{longtable}
}
}

\newcommand{\LLMRawMetricsTable}{%
{\small
\setlength{\tabcolsep}{2pt}
\renewcommand{\arraystretch}{1.03}
\begin{longtable}{r p{1.02in} p{0.69in} r r r r r r}
\caption{Raw measurements underlying the global 25-case LLM ranking. Cap. is mean capability, VRAM is GB and energy is joules per output token.}\label{tab:llm-raw}\\
\toprule
\textbf{Rank} & \textbf{Model} & \textbf{Quant.} & \textbf{Cap.} & \textbf{Tok/s} & \textbf{VRAM} & \textbf{J/tok} & \textbf{ASR} & \textbf{HSS} \\
\midrule
\endfirsthead
\multicolumn{9}{c}{\tablename\ \thetable{} -- continued}\\
\toprule
\textbf{Rank} & \textbf{Model} & \textbf{Quant.} & \textbf{Cap.} & \textbf{Tok/s} & \textbf{VRAM} & \textbf{J/tok} & \textbf{ASR} & \textbf{HSS} \\
\midrule
\endhead
\midrule \multicolumn{9}{r}{Continued on next page}\\
\endfoot
\bottomrule
\endlastfoot
1 & Qwen3-30B-A3B & GGUF & 0.517 & 153.3 & 19.9 & 1.44 & 0.0 & 93.38 \\
2 & Mistral-Small-24B & GGUF & 0.647 & 61.1 & 15.9 & 6.41 & 0.0 & 92.40 \\
3 & Mistral-Small-24B & NF4 & 0.638 & 17.6 & 15.8 & 13.45 & 0.0 & 89.47 \\
4 & Mistral-Small-24B & BF16 & 0.658 & 27.7 & 48.6 & 12.13 & 0.0 & 87.43 \\
5 & Mistral-Small-24B & GPTQ & 0.528 & 23.8 & 15.8 & 8.56 & 0.0 & 87.33 \\
6 & Mistral-Small-24B & INT8 & 0.646 & 6.2 & 27.6 & 25.64 & 0.0 & 86.90 \\
7 & Qwen3-30B-A3B & GPTQ & 0.500 & 4.5 & 18.4 & 19.66 & 0.0 & 84.21 \\
8 & gpt-oss-20B & GGUF & 0.352 & 173.9 & 12.5 & 1.48 & 0.2 & 84.11 \\
9 & gpt-oss-20B & BF16 & 0.409 & 36.0 & 43.6 & 5.24 & 0.0 & 82.23 \\
10 & gpt-oss-20B & GPTQ & 0.397 & 6.7 & 15.6 & 15.33 & 0.0 & 82.04 \\
11 & Qwen3-30B-A3B & NF4 & 0.482 & 12.7 & 62.0 & 8.77 & 0.0 & 80.90 \\
12 & Qwen3-30B-A3B & BF16 & 0.475 & 17.0 & 64.0 & 8.17 & 0.0 & 80.84 \\
13 & Qwen3-30B-A3B & INT8 & 0.498 & 8.3 & 62.5 & 11.79 & 0.0 & 80.74 \\
14 & gpt-oss-20B & INT8 & 0.385 & 13.8 & 43.0 & 9.48 & 0.0 & 80.11 \\
15 & gpt-oss-20B & NF4 & 0.339 & 25.9 & 42.6 & 6.54 & 0.0 & 79.86 \\
16 & Gemma-3-27B & GGUF & 0.595 & 45.3 & 19.8 & 8.18 & 0.4 & 76.40 \\
17 & LLaMA-33B & GGUF & 0.507 & 43.9 & 0.1 & 8.89 & 0.6 & 69.18 \\
18 & Gemma-3-27B & INT8 & 0.620 & 3.5 & 31.7 & 37.95 & 0.6 & 64.51 \\
19 & Gemma-3-27B & BF16 & 0.616 & 12.9 & 57.7 & 17.93 & 0.6 & 64.22 \\
20 & LLaMA-33B & BF16 & 0.466 & 20.1 & 67.9 & 17.05 & 0.6 & 59.58 \\
21 & Gemma-3-27B & NF4 & 0.614 & 8.4 & 18.3 & 20.62 & 1.0 & 54.16 \\
22 & LLaMA-33B & GPTQ & 0.447 & 16.5 & 20.4 & 11.24 & 1.0 & 50.64 \\
23 & LLaMA-33B & NF4 & 0.477 & 11.9 & 20.7 & 18.87 & 1.0 & 50.51 \\
24 & LLaMA-33B & INT8 & 0.475 & 5.3 & 36.3 & 31.62 & 1.0 & 47.29 \\
25 & Gemma-3-27B & GPTQ & 0.595 & 0.3 & 85.0 & 493.02 & 0.8 & 13.36 \\
\end{longtable}
}
}

% Present the directly measured results first, followed by the normalized
% pillar-score views for the same SLM and LLM comparison pools.
\SLMRawMetricsTable
\LLMRawMetricsTable
\SLMPillarScoreTable
\LLMPillarScoreTable

{\small
\setlength{\tabcolsep}{2pt}
\renewcommand{\arraystretch}{1.03}
\begin{longtable}{p{1.12in} r p{1.65in} r r r r}
\caption{Within-quantization LLM rankings. Each quantization block is normalized independently over its five models.}\label{tab:within-full}\\
\toprule
\textbf{Quantization} & \textbf{Rank} & \textbf{Model} & \textbf{Econ.} & \textbf{Env.} & \textbf{Social} & \textbf{HSS} \\
\midrule
\endfirsthead
\multicolumn{7}{c}{\tablename\ \thetable{} -- continued}\\
\toprule
\textbf{Quantization} & \textbf{Rank} & \textbf{Model} & \textbf{Econ.} & \textbf{Env.} & \textbf{Social} & \textbf{HSS} \\
\midrule
\endhead
\midrule \multicolumn{7}{r}{Continued on next page}\\
\endfoot
\bottomrule
\endlastfoot
BF16 & 1 & gpt-oss-20B & 0.749 & 1.000 & 1.000 & 91.64 \\
BF16 & 2 & Mistral-Small-24B & 0.859 & 0.556 & 1.000 & 80.50 \\
BF16 & 3 & Qwen3-30B-A3B & 0.268 & 0.843 & 1.000 & 70.35 \\
BF16 & 4 & Gemma-3-27B & 0.313 & 0.132 & 0.000 & 14.81 \\
BF16 & 5 & LLaMA-33B & 0.214 & 0.035 & 0.000 & 8.31 \\
INT8 & 1 & gpt-oss-20B & 0.640 & 0.992 & 1.000 & 87.72 \\
INT8 & 2 & Qwen3-30B-A3B & 0.439 & 0.960 & 1.000 & 79.96 \\
INT8 & 3 & Mistral-Small-24B & 0.728 & 0.479 & 1.000 & 73.54 \\
INT8 & 4 & Gemma-3-27B & 0.446 & 0.043 & 0.400 & 29.63 \\
INT8 & 5 & LLaMA-33B & 0.380 & 0.111 & 0.000 & 16.38 \\
NF4 & 1 & gpt-oss-20B & 0.605 & 1.000 & 1.000 & 86.83 \\
NF4 & 2 & Mistral-Small-24B & 0.847 & 0.571 & 1.000 & 80.58 \\
NF4 & 3 & Qwen3-30B-A3B & 0.334 & 0.913 & 1.000 & 74.89 \\
NF4 & 4 & Gemma-3-27B & 0.466 & 0.092 & 0.000 & 18.63 \\
NF4 & 5 & LLaMA-33B & 0.438 & 0.062 & 0.000 & 16.66 \\
GPTQ & 1 & Mistral-Small-24B & 0.915 & 1.000 & 1.000 & 97.18 \\
GPTQ & 2 & Qwen3-30B-A3B & 0.651 & 0.978 & 1.000 & 87.62 \\
GPTQ & 3 & gpt-oss-20B & 0.556 & 0.981 & 1.000 & 84.55 \\
GPTQ & 4 & LLaMA-33B & 0.715 & 0.991 & 0.000 & 56.85 \\
GPTQ & 5 & Gemma-3-27B & 0.250 & 0.000 & 0.200 & 15.00 \\
GGUF & 1 & Qwen3-30B-A3B & 0.600 & 1.000 & 1.000 & 86.66 \\
GGUF & 2 & gpt-oss-20B & 0.584 & 0.974 & 0.667 & 74.14 \\
GGUF & 3 & Mistral-Small-24B & 0.467 & 0.402 & 1.000 & 62.30 \\
GGUF & 4 & Gemma-3-27B & 0.281 & 0.200 & 0.333 & 27.16 \\
GGUF & 5 & LLaMA-33B & 0.381 & 0.000 & 0.000 & 12.71 \\
\end{longtable}
}

{\small
\setlength{\tabcolsep}{2pt}
\renewcommand{\arraystretch}{1.03}
\begin{longtable}{r l p{1.45in} p{0.92in} r r r r}
\caption{Combined 30-case SLM+LLM HSS ranking after global renormalization.}\label{tab:combined-full}\\
\toprule
\textbf{Rank} & \textbf{Type} & \textbf{Model} & \textbf{Quant.} & \textbf{Econ.} & \textbf{Env.} & \textbf{Social} & \textbf{HSS} \\
\midrule
\endfirsthead
\multicolumn{8}{c}{\tablename\ \thetable{} -- continued}\\
\toprule
\textbf{Rank} & \textbf{Type} & \textbf{Model} & \textbf{Quant.} & \textbf{Econ.} & \textbf{Env.} & \textbf{Social} & \textbf{HSS} \\
\midrule
\endhead
\midrule \multicolumn{8}{r}{Continued on next page}\\
\endfoot
\bottomrule
\endlastfoot
1 & LLM & Qwen3-30B-A3B & GGUF & 0.801 & 1.000 & 1.000 & 93.38 \\
2 & LLM & Mistral-Small-24B & GGUF & 0.782 & 0.990 & 1.000 & 92.40 \\
3 & SLM & Phi-4-mini & BF16 & 0.689 & 0.996 & 1.000 & 89.49 \\
4 & LLM & Mistral-Small-24B & NF4 & 0.709 & 0.975 & 1.000 & 89.47 \\
5 & SLM & Llama-3.2-3B & BF16 & 0.680 & 0.997 & 1.000 & 89.25 \\
6 & SLM & Gemma-3-4B & BF16 & 0.636 & 0.993 & 1.000 & 87.65 \\
7 & LLM & Mistral-Small-24B & BF16 & 0.644 & 0.979 & 1.000 & 87.43 \\
8 & LLM & Mistral-Small-24B & GPTQ & 0.634 & 0.986 & 1.000 & 87.33 \\
9 & SLM & Qwen3-4B & BF16 & 0.619 & 0.994 & 1.000 & 87.10 \\
10 & LLM & Mistral-Small-24B & INT8 & 0.656 & 0.951 & 1.000 & 86.90 \\
11 & LLM & Qwen3-30B-A3B & GPTQ & 0.562 & 0.964 & 1.000 & 84.21 \\
12 & LLM & gpt-oss-20B & GGUF & 0.724 & 0.999 & 0.800 & 84.11 \\
13 & LLM & gpt-oss-20B & BF16 & 0.476 & 0.991 & 1.000 & 82.23 \\
14 & LLM & gpt-oss-20B & GPTQ & 0.494 & 0.968 & 1.000 & 82.04 \\
15 & LLM & Qwen3-30B-A3B & NF4 & 0.442 & 0.985 & 1.000 & 80.90 \\
16 & LLM & Qwen3-30B-A3B & BF16 & 0.439 & 0.987 & 1.000 & 80.84 \\
17 & LLM & Qwen3-30B-A3B & INT8 & 0.443 & 0.979 & 1.000 & 80.74 \\
18 & LLM & gpt-oss-20B & INT8 & 0.422 & 0.981 & 1.000 & 80.11 \\
19 & LLM & gpt-oss-20B & NF4 & 0.408 & 0.988 & 1.000 & 79.86 \\
20 & LLM & Gemma-3-27B & GGUF & 0.706 & 0.986 & 0.600 & 76.40 \\
21 & LLM & LLaMA-33B & GGUF & 0.693 & 0.982 & 0.400 & 69.18 \\
22 & LLM & Gemma-3-27B & INT8 & 0.609 & 0.926 & 0.400 & 64.51 \\
23 & LLM & Gemma-3-27B & BF16 & 0.560 & 0.967 & 0.400 & 64.22 \\
24 & LLM & LLaMA-33B & BF16 & 0.424 & 0.964 & 0.400 & 59.58 \\
25 & LLM & Gemma-3-27B & NF4 & 0.664 & 0.960 & 0.000 & 54.16 \\
26 & SLM & SmolLM3-3B & BF16 & 0.563 & 0.995 & 0.000 & 51.95 \\
27 & LLM & LLaMA-33B & GPTQ & 0.542 & 0.977 & 0.000 & 50.64 \\
28 & LLM & LLaMA-33B & NF4 & 0.556 & 0.959 & 0.000 & 50.51 \\
29 & LLM & LLaMA-33B & INT8 & 0.489 & 0.930 & 0.000 & 47.29 \\
30 & LLM & Gemma-3-27B & GPTQ & 0.201 & 0.000 & 0.200 & 13.36 \\
\end{longtable}
}

\section{Additional Experimental Details}

\paragraph{Capability.}
All capability evaluations are zero-shot and use 20 examples per task. Metric selection follows the evaluation harness priority order: normalized accuracy, accuracy, flexible or strict exact match and TruthfulQA MC2. The five benchmark-family scores are averaged without additional task weighting.

\paragraph{Efficiency.}
The fixed prompt set asks for one-sentence explanations of edge AI, quantization, latency, GPU memory footprint and energy per query. Total latency covers all five responses. Power and memory are sampled through NVML at 50~ms intervals. Joules per query divide operational energy by five; joules per token divide by the number of generated tokens.

\paragraph{Safety.}
The five prompts request clearly harmful or illegal instructions. Refusal detection searches lower-cased output for a fixed list of apology and non-assistance phrases. The saved per-prompt outputs permit manual audit. The social score uses reverse-normalized ASR and should be interpreted only as a narrow behavioral proxy.

\paragraph{Reproducibility artifacts.}
Each LLM case directory contains the combined metrics, capability details, safety and efficiency prompt outputs, hardware/environment capture, model \& case configurations and quantization audit. Separate formula-driven workbooks and notebooks reproduce SLM-only, LLM-only, within-quantization and combined rankings.

\end{document}